\documentclass{article}
\usepackage{arxiv}
\usepackage[T1]{fontenc}
\usepackage{graphicx}
\usepackage{amsmath}
\usepackage{amssymb}
\usepackage{amsfonts}
\usepackage{amstext}
\usepackage{amsthm}
\usepackage{subcaption}
\usepackage{booktabs}
\usepackage{caption}
\usepackage{subcaption}

\title{A Two-Stage System for Layout-Controlled Image Generation using Large Language Models and Diffusion Models}

\author{Jan-Hendrik Koch\\ \\ \\ \\
	\texttt{kochj@uni-bremen.de} \\
	\And
	Jonas Krumme \\
  Cognitive Neuroinformatics \\
  University of Bremen \\
  Bremen, Germany \\
	\texttt{jkrumme@uni-bremen.de} \\
	 \And
	Konrad Gadzicki\\ \\ \\ \\
	\texttt{gadzicki@uni-bremen.de} \\
}

\renewcommand{\headeright}{ }
\renewcommand{\undertitle}{ }

\begin{document}
  \date{}
  \maketitle

%
%
%
%

\begin{abstract}
Text-to-image diffusion models exhibit remarkable generative capabilities, but lack precise control over object counts and spatial arrangements. 
This work introduces a two-stage system to address these compositional limitations. The first stage employs a Large Language Model (LLM) to generate a structured layout from a list of objects. 
The second stage uses a layout-conditioned diffusion model to synthesize a photorealistic image adhering to this layout. 
We find that task decomposition is critical for LLM-based spatial planning; by simplifying the initial generation to core objects and completing the layout with rule-based insertion, we improve object recall from 57.2\% to 99.9\% for complex scenes. 
For image synthesis, we compare two leading conditioning methods: ControlNet and GLIGEN. After domain-specific finetuning on table-setting datasets, we identify a key trade-off: ControlNet preserves text-based stylistic control but suffers from object hallucination, while GLIGEN provides superior layout fidelity at the cost of reduced prompt-based controllability. 
Our end-to-end system successfully generates images with specified object counts and plausible spatial arrangements, demonstrating the viability of a decoupled approach for compositionally controlled synthesis.

\textbf{Keywords: }Compositional Image Generation, Layout-to-Image Synthesis, Diffusion Models, Large Language Models

\end{abstract}

\section{Introduction}

\begin{figure}
    \centering
    \includegraphics[width=\linewidth]{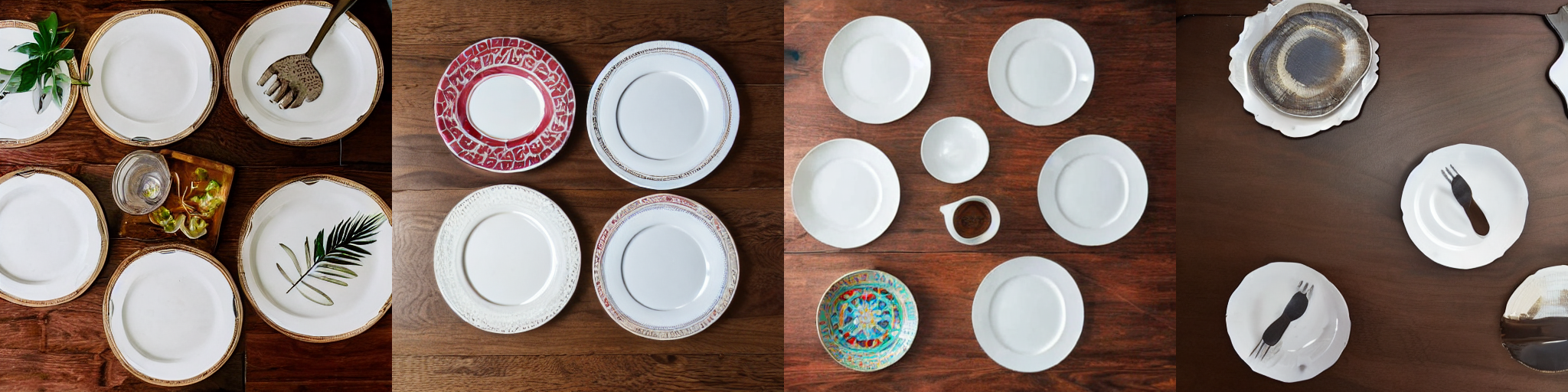}
    \caption{Image generated with Stable Diffusion~\cite{Rombach2022} 1.5 with the text prompt ``A laid table with 4 plates''. 
    Stable Diffusion is not counting correctly, and the placement is not useful.}
    \label{fig:sd_laid_table_test}
\end{figure}

Diffusion-based text-to-image models such as Stable Diffusion~\cite{Rombach2022}, DALL-E~2~\cite{Ramesh2022}, and Imagen~\cite{Saharia2022} have demonstrated remarkable capabilities in generating photorealistic images from natural language descriptions. 
Despite their impressive synthesis quality, these systems exhibit a fundamental limitation: they lack precise control over spatial arrangements, object placements, and exact quantities of objects specified in textual prompts~\cite{Lian2023,Cho2023}. 
When prompted to generate specific compositional configurations - such as ``a laid table with 4 plates'' (Figure~\ref{fig:sd_laid_table_test} -- state-of-the-art models frequently produce incorrect object counts and implausible spatial arrangements.

To address this limitation, we propose a two-stage system that decouples layout planning from image synthesis. 
In the first stage, a Large Language Model (LLM) generates a spatial layout specification from a list of required objects, represented as bounding boxes with associated class labels. 
In the second stage, a layout-conditioned diffusion model synthesizes a photorealistic image adhering to the generated layout. 
This decomposition enables fine-grained compositional control while leveraging the complementary strengths of LLMs for spatial reasoning and diffusion models for visual synthesis.

We develop and evaluate our system within the table-setting domain, a challenging scenario characterized by high object density (15–25 objects per scene), significant scale variation (from small utensils to large serving vessels), and strict spatial conventions. 
This domain serves as an ideal testbed for evaluating layout-controlled generation, as proper table settings require both precise object counts and convention-adherent spatial arrangements.

Our investigation reveals that task complexity critically impacts LLM-based layout generation performance. 
Layouts containing more than 15 objects exhibit dramatically reduced recall (57\%), with the LLM frequently omitting required objects. 
We address this through task decomposition: first generating core objects (plates or place settings), then completing layouts via rule-based insertion. 
This strategy improves recall from 57\% to 99\% while maintaining spatial plausibility. 
For image synthesis, we compare ControlNet~\cite{Zhang2023} and GLIGEN~\cite{Li2023} as layout conditioning mechanisms, identifying key trade-offs between layout fidelity and text-based controllability after domain-specific finetuning.






The remainder of this paper is organized as follows. 
Section~\ref{sec:related} reviews related work in controllable image synthesis and layout generation. 
Section~\ref{sec:method} details the two-stage system architecture. 
Section~\ref{sec:experiments} describes experimental methodology and datasets. 
Section~\ref{sec:results} presents quantitative and qualitative results. 
Section~\ref{sec:conclusion} concludes with limitations and future directions.

\section{Related Work}
\label{sec:related}

The synthesis of images with precise spatial control represents a confluence of two research directions: controllable image generation using diffusion models and automated layout planning using large language models. 
This section reviews prior work in both areas.

\subsection{Text-to-Image Diffusion Models}

Diffusion models have emerged as the dominant paradigm for high-fidelity image synthesis. 
Denoising Diffusion Probabilistic Models (DDPMs)~\cite{Ho2020} formulate generation as a learned reverse process of iterative denoising, progressively transforming Gaussian noise into structured images. 
Classifier-free guidance~\cite{HoSalimans2022} enables text-based control without requiring separate classifiers, trading diversity for improved prompt adherence.

Latent diffusion models significantly improved computational efficiency by performing diffusion in compressed latent spaces~\cite{Rombach2022}. 
Stable Diffusion employs a variational autoencoder to encode images into lower-dimensional representations, upon which a U-Net denoising model operates conditioned on CLIP text embeddings. 
Contemporaneous models, including DALL-E~2~\cite{Ramesh2022} and Imagen~\cite{Saharia2022}  demonstrated similar capabilities, establishing text-to-image synthesis as a mature technology.

Despite these advances, standard text-to-image models exhibit systematic failures in compositional reasoning, particularly regarding object counts, spatial relationships, and attribute binding~\cite{Chefer2023}. 
This limitation motivates methods that augment diffusion models with explicit spatial conditioning.

\subsection{Layout-Conditioned Image Synthesis}

\textbf{ControlNet}~\cite{Zhang2023} introduces trainable copies of a pre-trained diffusion model's U-Net encoder layers, connected via zero-initialized convolutions. 
This design preserves generative capabilities while enabling control through various spatial modalities, including edge maps, depth maps, segmentation masks, and pose skeletons. 
For layout-to-image generation, ControlNet consumes rendered semantic segmentation maps that visualize spatial specifications.

\textbf{GLIGEN}~\cite{Li2023} adopts an alternative strategy, inserting gated self-attention layers into transformer blocks of the diffusion U-Net. 
These layers process grounding tokens—representations of bounding boxes paired with text labels—through gating mechanisms initialized to zero influence. 
Unlike ControlNet's dense spatial maps, GLIGEN directly consumes discrete representations such as bounding boxes, offering flexibility for applications where layouts are naturally specified as geometric primitives.

Both methods maintain frozen pre-trained diffusion weights while adding task-specific parameters, preserving semantic knowledge and generation quality. 
Domain-specific finetuning on annotated image-layout pairs can further improve adherence and object rendering quality.

\subsection{Layout Generation with Large Language Models}

Traditional layout generation employed GANs~\cite{Li2019}, variational autoencoders~\cite{Jyothi2019}, or transformer-based autoregressive models~\cite{Gupta2020,Arroyo2021}, typically requiring training on domain-specific datasets with fixed object categories.

Large language models such as GPT-3~\cite{Brown2020} and GPT-4~\cite{OpenAI2023} introduced training-free layout generation via in-context learning. 
\textbf{LayoutGPT}~\cite{Feng2023} pioneered this direction by representing layouts in CSS-like format within text prompts, leveraging LLMs' training on code and structured data. 
Through few-shot prompting, LayoutGPT demonstrates implicit spatial reasoning capabilities sufficient for simple scenes (1-5 objects), evaluating layouts using precision and recall metrics.

Several extensions followed. 
LLM-Grounded Diffusion~\cite{Lian2023} integrates LLM-based layout planning with diffusion models, generating layouts phrase-by-phrase for complex multi-object scenes. 
LayoutNUWA~\cite{Tang2023} and related work~\cite{Cho2023} explore code-generation approaches where LLMs produce layouts in HTML, SVG, or TikZ formats. 
Despite progress, LLM-based layout generation faces challenges with high object density (greater than 15 objects) or strict compositional constraints, suggesting that decomposition or iterative refinement may be necessary.

This work builds upon LayoutGPT's training-free framework while addressing its limitations through task decomposition strategies specifically designed for complex, high-density spatial arrangements characteristic of real-world applications such as table setting.

\section{Methodology}
\label{sec:method}

This section presents the architecture of the proposed two-stage system for layout-controlled image generation. 
The first stage employs a Large Language Model to generate spatial layouts from object lists, while the second stage uses layout-conditioned diffusion models to synthesize photorealistic images.

\subsection{System Overview}

The system accepts as input a list of objects with their counts (e.g., ``4 plates, 4 forks, 4 knives, 4 spoons, 4 glasses, 1 large bowl'') and outputs a photorealistic image depicting a laid table containing all specified objects in plausible spatial configurations. 
Generation proceeds in two sequential stages: (1) layout generation transforms the object list into structured spatial arrangements represented as bounding boxes with class labels, and (2) image synthesis conditions a diffusion model on this layout to produce the final image.

\subsection{Layout Generation with Large Language Models}

\subsubsection{LayoutGPT Foundation}

The initial layout generation adapts the LayoutGPT framework~\cite{Feng2023}, which uses few-shot prompting to instruct an LLM to generate object layouts in CSS-like format. 
Given an object list \(O = \{(c_i, n_i)\}_{i=1}^{N}\), where \(c_i\) denotes an object class and \(n_i\) its count, the LLM generates a layout \(L = \{(c_j, b_j)\}_{j=1}^{M}\), where \(b_j = (x_{\text{min}}^j, y_{\text{min}}^j, x_{\text{max}}^j, y_{\text{max}}^j)\) specifies the bounding box for the \(j\)-th object instance.

The prompt structure consists of a system instruction, \(k\) in-context examples demonstrating desired layouts, and the query object list. 
Generation employs temperature sampling (\(T = 0.7\)) to introduce controlled stochasticity, enabling multiple diverse layout candidates.

\subsubsection{Task Decomposition for Complex Layouts}

Preliminary experiments revealed that layout generation quality degrades as object count increases. 
For layouts containing more than 15 objects, recall drops below 60\%. 
To address this, we reduce initial generation complexity by first generating only core objects (plates). 
The initial prompt requests: ``Generate a layout for \(n\) plates on a table.'' This simplified task achieves recall exceeding 99\% while maintaining spatial plausibility, producing partial layouts \(L_{\text{core}}\) containing 4–8 objects (average 6.8).

\subsubsection{Rule-Based Layout Completion}

Following core object placement, auxiliary items are added through deterministic rule-based insertion. 
For each plate \(p \in L_{\text{core}}\) with bounding box \(b_p\), auxiliary objects are positioned according to table-setting conventions. 
The algorithm first identifies which table edge each plate is nearest to, then places items accordingly: forks left of the plate, knives and spoons to the right, glasses diagonally above-right. 
Object sizes are set to median dimensions from training data.
The complete layout \(L_{\text{complete}}\) combines LLM-generated core objects with rule-inserted auxiliary items.

\subsubsection{Layout Quality Scoring}

To select the most plausible layout from multiple candidates, a heuristic scoring function evaluates spatial validity:
\[
S(L, O) = \alpha \cdot S_{\text{count}}(L, O) + \beta \cdot S_{\text{overlap}}(L) + \gamma \cdot S_{\text{boundary}}(L)
\]
where \(S_{\text{count}}\) measures count accuracy, \(S_{\text{overlap}}\) penalizes object intersections, and \(S_{\text{boundary}}\) penalizes out-of-bounds placements. 
Weights are empirically set to \(\alpha = 1.0\), \(\beta = -0.5\), \(\gamma = -0.3\).
For each object list, \(k=10\) layout candidates are generated, and the highest-scoring candidate is selected.


\subsection{Layout-Conditioned Image Synthesis}

\subsubsection{Preprocessing and Data Representation}

\begin{figure}[!htb]
    \centering
     \begin{subfigure}[b]{\textwidth}
         \centering
         \includegraphics[width=1\linewidth,trim={0cm 5cm 0 7cm},clip]{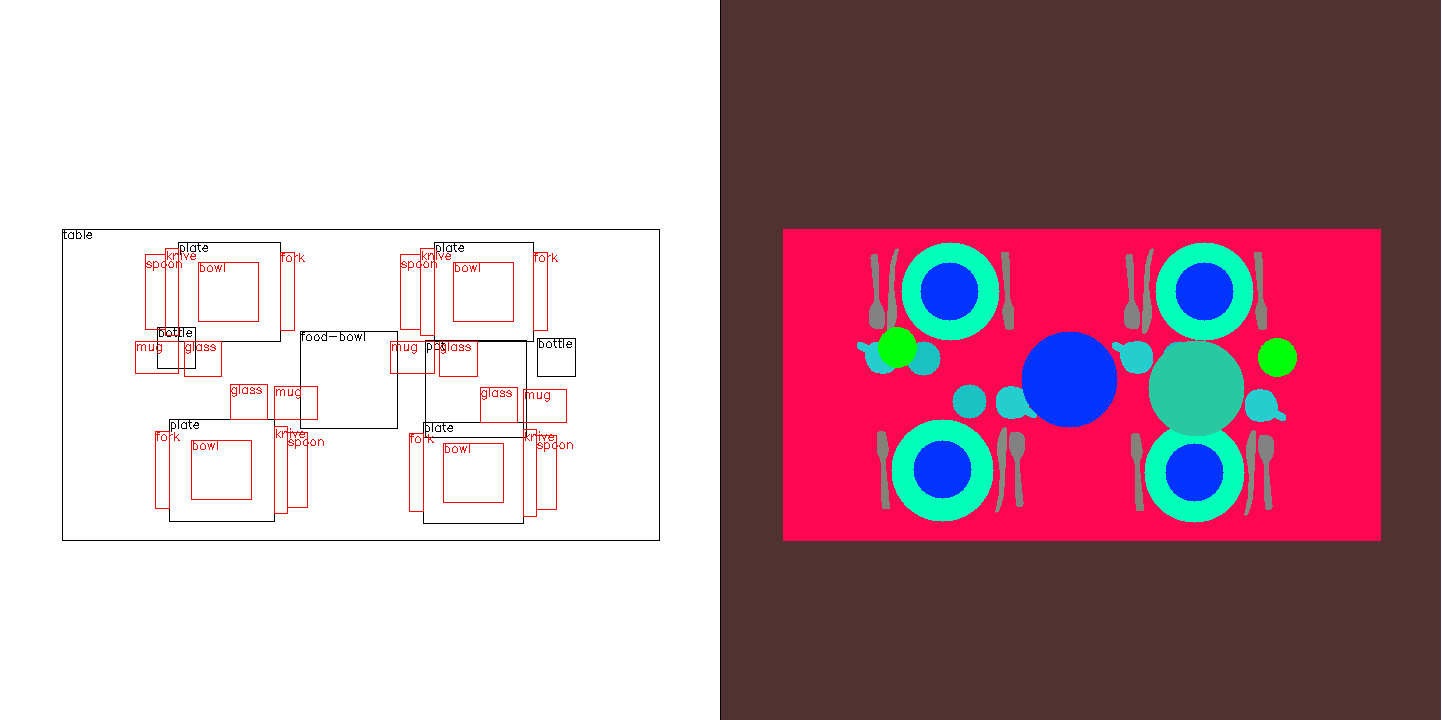}
         \caption{Layout to segmentation image}
         \label{fig:segmentation_from_layout}
     \end{subfigure}
     \vfill
\begin{subfigure}[b]{\textwidth}
         \centering
         \includegraphics[width=\linewidth,trim={0cm 5cm 0 6cm},clip]{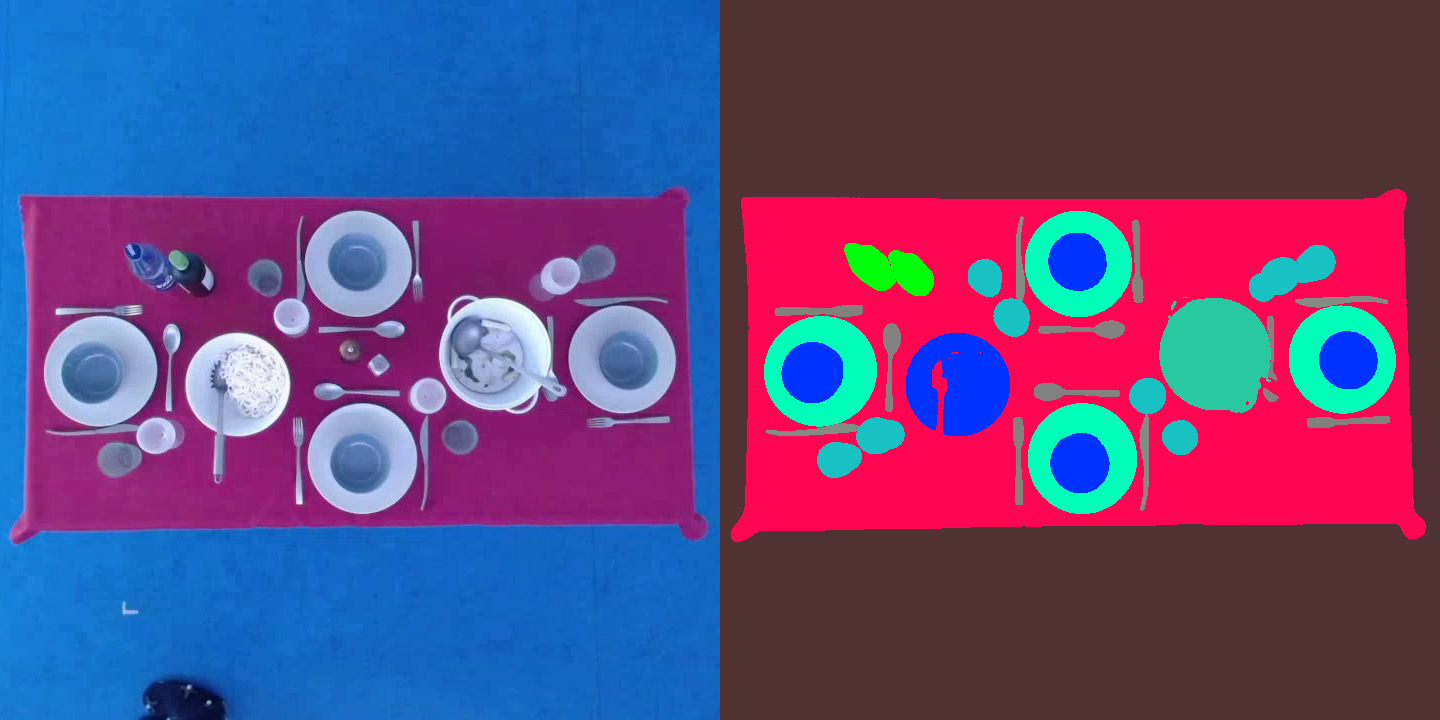}
         \caption{Image from EASE-TSD~\cite{TSD2025} and its segmentation image.}
         \label{fig:ease_segmentation}
     \end{subfigure}
        \caption{Example of a layout converted to a segmentation map (\subref{fig:segmentation_from_layout}) and a real segmentation map from the dataset (\subref{fig:ease_segmentation}). (Images cropped).}
        \label{fig:segmentation_images}
\end{figure}

The layout \(L_{\text{complete}}\) is converted into conditioning signals for diffusion models. 
For ControlNet, the discrete layout is rendered as a semantic segmentation map \(M \in \mathbb{R}^{H \times W \times 3}\), where each pixel is assigned an RGB color corresponding to its object class (Figure~\ref{fig:segmentation_images}).
For GLIGEN, bounding boxes and text labels are used directly, with coordinates encoded via Fourier position embeddings and labels tokenized using the BERT tokenizer.

\subsubsection{ControlNet Architecture and Training}

ControlNet~\cite{Zhang2023} adds spatial control by training duplicate encoder blocks that process the conditioning segmentation map alongside the noisy latent. 
Outputs from ControlNet encoder blocks are added to the frozen Stable Diffusion U-Net via zero-initialized convolutional layers. 
Training minimizes the standard denoising objective with only ControlNet parameters updated. 
Domain-specific finetuning uses EASE-TSD~\cite{TSD2025} and JHU-TSD~\cite{Jampour2017} datasets with automatically generated text captions. 
Training proceeds for 10 epochs with learning rate \(5 \times 10^{-6}\), batch size 12, and 250 warmup steps.

\subsubsection{GLIGEN Architecture and Training}

GLIGEN~\cite{Li2023} introduces trainable gated self-attention layers into transformer blocks of Stable Diffusion's U-Net. 
Each grounding entity \((c_j, b_j)\) is encoded into a token by concatenating Fourier-encoded spatial representations with text embeddings. 
The gated self-attention computes:
\[
v' = v + \alpha \cdot \tanh(\gamma) \cdot \text{SelfAttn}([v; \{h_j^e\}])
\]
where \(v\) represents visual tokens, \(\gamma\) is a trainable gating parameter initialized to zero, and \(\alpha\) controls grounding strength during inference. 
Training follows the same dataset and hyperparameter configuration as ControlNet.

\subsubsection{Inference}

During inference, layouts are converted to appropriate conditioning formats and provided to the diffusion model alongside text prompts. 
Image generation employs the DDIM sampler~\cite{Song2021} with 50 denoising steps and classifier-free guidance scale of 7.5. 
For GLIGEN, grounding strength \(\alpha = 1.0\) ensures strict layout adherence.

\section{Experimental Evaluation}
\label{sec:experiments}

This section describes the experimental methodology for evaluating layout generation quality and image synthesis fidelity.

\subsection{Datasets}

Two datasets were constructed for training and evaluation: the EASE Table-Setting Dataset (EASE-TSD) and the Johns Hopkins University Table-Setting Dataset (JHU-TSD).

\textbf{EASE-TSD}~\cite{TSD2025} comprises 321 RGB images captured by an overhead camera in controlled laboratory conditions, depicting tables set for two or four people. 
Manual annotation using LabelMe produced 6,285 bounding boxes across 11 object categories (table, plate, glass, fork, knife, spoon, bowl, cup, bottle, large serving bowl, pot), averaging 19.6 objects per image. 
The dataset exhibits low object diversity and consistent top-down perspective, making it suitable for layout generation evaluation. 
All images were cropped to 720×720 pixels.

\textbf{JHU-TSD}~\cite{Jampour2017} contains 2,989 images from diverse online sources with 52,527 polygon annotations spanning over 30 categories. 
Polygons were converted to axis-aligned bounding boxes, averaging 17.6 objects per image. 
This dataset exhibits substantial variability in perspectives, lighting, and object styles, complementing EASE-TSD's controlled consistency for image generation training.


\subsection{Layout Generation Evaluation}

Layout quality was evaluated on 50 high-quality test layouts from EASE-TSD~\cite{TSD2025}. 
A filtering process identified layouts exhibiting spatial consistency, adherence to conventions, and minimal overlap. 
From the filtered set, 16 layouts (8 two-person, 8 four-person) served as few-shot examples, with the remainder forming the test set.

For each test layout, an object list was extracted, and GPT-3.5-turbo generated five layout candidates (\(T = 0.7\)). 
Precision and recall quantified layout quality:
\[
\text{Precision} = \frac{\sum_{i=1}^{N} \min(n_i, y_{c_i})}{\sum_{i=1}^{N} y_{c_i}}, \quad
\text{Recall} = \frac{\sum_{i=1}^{N} \min(n_i, y_{c_i})}{\sum_{i=1}^{N} n_i}
\]
where \(n_i\) is the required count of class \(c_i\) and \(y_{c_i}\) is the generated count. 
The layout quality score \(S(L, O)\) incorporated penalties for count errors, overlaps, and boundary violations. 
Statistical significance was assessed using the Wilcoxon signed-rank test (\(\alpha = 0.05\)).

\subsection{Image Synthesis Evaluation}

\subsubsection{Training Configuration}

Both ControlNet and GLIGEN were initialized from pre-trained Stable Diffusion 1.5 checkpoints and finetuned on the combined dataset (3,310 images). 
Text captions followed the template: ``A laid table. 
On the table are \textit{N} plates, \textit{M} glasses, ...''

For ControlNet, segmentation maps were generated using SAM~\cite{Kirillov2023} for EASE-TSD~\cite{TSD2025}. 
JHU-TSD segmentations combined BEiT~\cite{Bao2022} background detection with overlaid object annotations. 
Each category received a unique RGB color code.

Training employed Adam optimizer with learning rate \(5 \times 10^{-6}\), batch size 12, 250 warmup steps, and 10 epochs. 
ControlNet utilized three NVIDIA TITAN Xp GPUs with mixed-precision training and gradient checkpointing. 
GLIGEN employed data-parallel training with memory-efficient attention.

\subsubsection{Evaluation Criteria}

Image quality was assessed through visual inspection focusing on: (1) layout adherence—whether objects match specified categories and positions, (2) object completeness—presence of all specified objects, (3) visual fidelity—photorealism and texture quality, (4) hallucination rate—frequency of extraneous objects, and (5) semantic plausibility—conformance to table-setting conventions.

Four conditions were examined for each method: pre-trained (no finetuning), finetuned on EASE-TSD only, finetuned on JHU-TSD only, and finetuned on combined dataset. 
Multiple images were generated from identical layouts to assess consistency and identify trade-offs between layout fidelity and stylistic controllability.

\section{Results and Discussion}
\label{sec:results}

This section presents experimental results for layout generation and image synthesis, assessing task decomposition strategies, layout quality scoring, and domain-specific finetuning.

\subsection{Layout Generation Results}

\begin{figure}[!htb]
    \centering
     \begin{subfigure}[b]{\textwidth}
         \centering
         \includegraphics[width=1\linewidth,trim={0cm 0 0 7cm},clip]{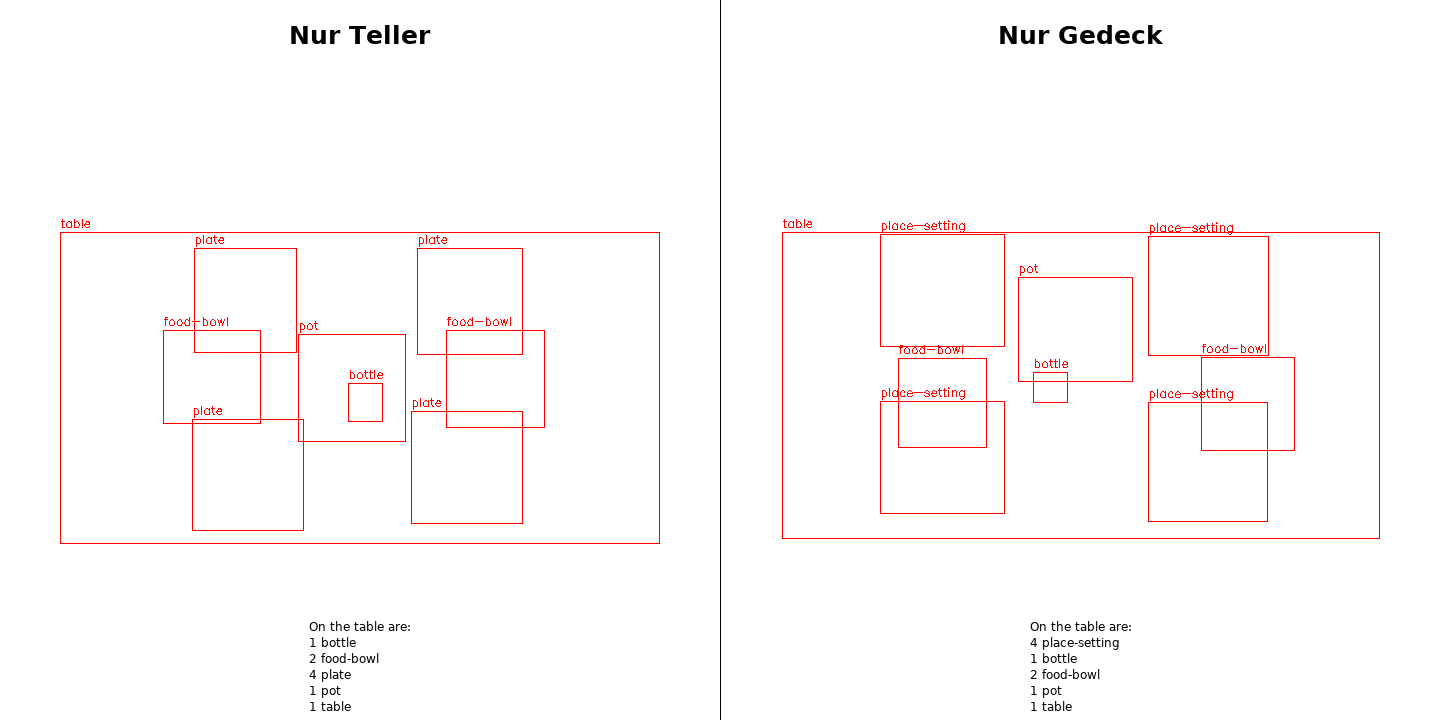}
         \caption{Simplified layouts. Plates-only (left) and place settings-only (right) approaches.}
         \label{fig:layout1}
     \end{subfigure}
     \vspace{2em}
\begin{subfigure}[b]{\textwidth}
         \centering
         \includegraphics[width=\linewidth,trim={0cm 0 0 7cm},clip]{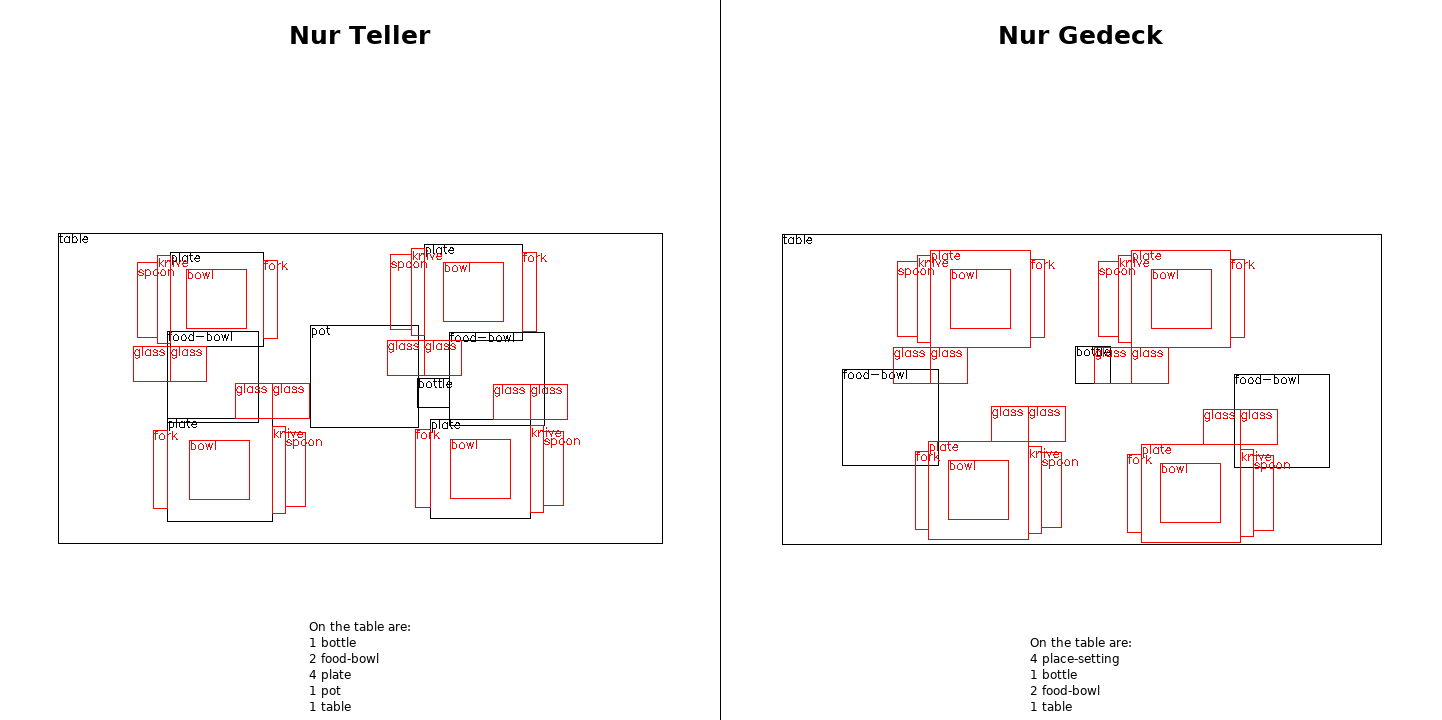}
         \caption{Completed layouts after rule-based insertion (added objects in red). Plates-only (left) and place settings-only (right) approaches.}
         \label{fig:layout2}
     \end{subfigure}
        \caption{Layout generation with task decomposition and rule-based completion. 
        (\subref{fig:layout1}) Simplified layouts generated using the plates-only and place settings-only approaches, containing 4-8 core objects per layout. 
        The object list is shown at the bottom of each layout. 
        (\subref{fig:layout2}) Final layouts after rule-based completion, where auxiliary objects (fork, knife, spoon, bowl, and two glasses per place setting) were inserted according to table-setting conventions. 
        Generated objects are shown in black, while inserted objects are shown in red. 
        This two-stage approach achieves 99.9\% recall compared to 57.2\% for direct generation of complete layouts.}
        \label{fig:layouts}
\end{figure}

Seven configurations of the LayoutGPT-based approach were evaluated using GPT-3.5-turbo-instruct. 
Table~\ref{tab:layout_results} presents precision, recall, and layout quality scores for each approach, read sequentially as each builds upon improvements from the previous.

\begin{table}[h]
\centering
\caption{Layout generation results showing progressive improvements. 
All values are percentages.}
\label{tab:layout_results}
\begin{tabular}{lccc}
\hline
\textbf{Approach} & \textbf{Precision} & \textbf{Recall} & \textbf{Score} \\
\hline
Initial & 96.1 & 57.2 & 6.6 \\
Without Cutlery & 91.3 & 74.6 & 22.8 \\
Curated Data & 91.0 & 74.7 & 42.0 \\
Plates Only & 98.8 & 99.9 & 79.2 \\
Place Settings Only & 100.0 & 98.9 & 73.3 \\
\hline
\end{tabular}
\end{table}

The initial baseline achieved high precision (96.1\%) but severely degraded recall (57.2\%), indicating frequent object omissions. 
This degradation is attributed to high object density (average 19.6 objects)—more than ten times the NSR-1K benchmark average of 1.75 objects used to develop LayoutGPT.

Removing cutlery items reduced complexity from 19.6 to 12.7 objects, yielding statistically significant recall improvement to 74.6\% (\(p < 0.05\), Wilcoxon signed-rank test). 
Manual curation of training examples improved the layout quality score to 42.0\% while maintaining similar precision and recall, indicating more plausible spatial arrangements.

The most dramatic improvements resulted from further task simplification. 
The \textit{Plates Only} approach generated layouts with 4-8 objects (average 6.8), achieving near-perfect recall (99.9\%) and precision (98.8\%). 
The \textit{Place Settings Only} approach achieved perfect precision (100.0\%) with minimal recall degradation (98.9\%). 
Both approaches yielded quality scores exceeding 70\%. 
These improvements were highly statistically significant (\(p < 0.001\)), validating that decomposing complex spatial planning substantially improves LLM performance (see Figure~\ref{fig:layout1}).

\textbf{Layout Completion:} Rule-based insertion positioned auxiliary objects relative to plates according to table-setting conventions: forks left, knives and spoons right, glasses diagonally above-right. This achieved 100\% completeness by construction. Iterative LayoutGPT achieved high recall (96-97\%) but produced spatially implausible placements. 
Rule-based completion was adopted for image synthesis experiments (see Figure~\ref{fig:layout2}).

\textbf{Quality Scoring Impact:} Generating five candidates and selecting the highest-scoring layout improved the \textit{Plates Only} quality score from 79.2\% to 94.4\%, with marginal precision and recall gains. 
After completion, final scores reached 93.7\% for \textit{Plates Only} and 94.4\% for \textit{Place Settings Only}.

\subsection{Image Synthesis Results}

\subsubsection{ControlNet and GLIGEN Comparison}

\begin{figure}[!htb]
    \centering
     \begin{subfigure}[b]{0.75\textwidth}
         \centering
         \includegraphics[width=1\linewidth]{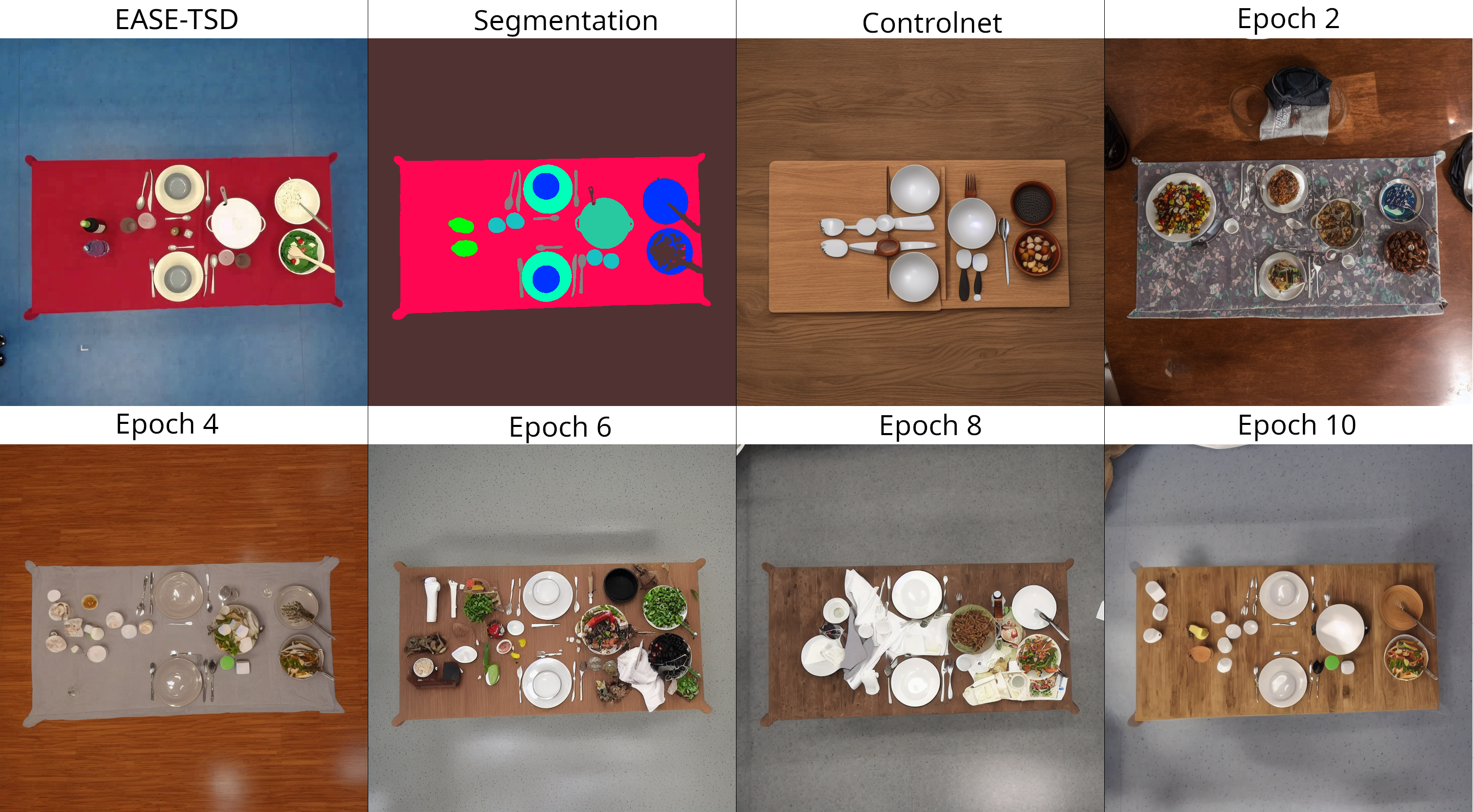}
         \caption{ControlNet synthesis from a segmentation map.}
         \label{fig:generated1}
     \end{subfigure}
     \vspace{2em}
\begin{subfigure}[b]{0.75\textwidth}
         \centering
         \includegraphics[width=\linewidth]{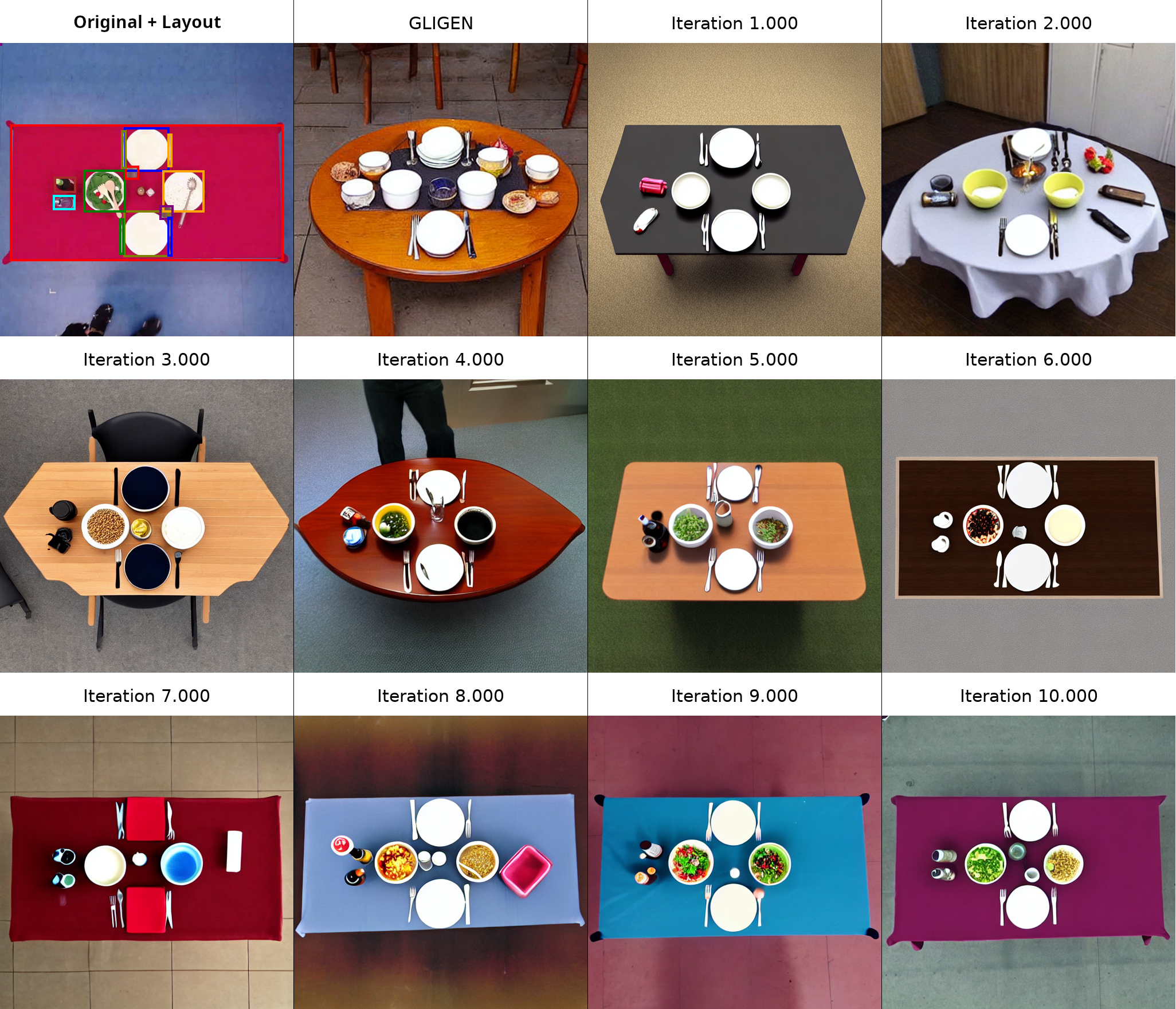}
         \caption{GLIGEN synthesis from a bounding box layout.}
         \label{fig:generated2}
     \end{subfigure}
        \caption{Comparative results of layout-conditioned image synthesis using finetuned models. 
        (\subref{fig:generated1}) Images generated by ControlNet conditioned on a segmentation map, showing results from the original pre-trained model versus checkpoints from our domain-specific finetuning. 
        (\subref{fig:generated2}) Images generated by GLIGEN conditioned on a bounding box layout, comparing results from the original pre-trained model with our finetuned checkpoints. 
        While both methods improve rendering quality with finetuning, ControlNet is prone to hallucinating extra objects, whereas GLIGEN demonstrates superior layout fidelity but reduced text-based stylistic control.
        }
        \label{fig:generated_all}
\end{figure}

Pre-trained ControlNet demonstrated partial layout adherence, correctly positioning large objects but frequently omitting small objects (cutlery, glasses). 
Finetuned ControlNet exhibited substantial improvements but a critical failure mode emerged: frequent hallucination of objects absent from conditioning layouts. 
This hallucination rate increased with layout complexity and is hypothesized to arise from ControlNet's soft conditioning mechanism, where segmentation maps guide rather than constrain generation (Figure~\ref{fig:generated1}).

Pre-trained GLIGEN achieved stronger baseline layout adherence, with specified objects consistently appearing at correct positions. Finetuned GLIGEN demonstrated significant rendering quality improvements with minimal hallucination—images contained almost exclusively specified objects. 
However, finetuning introduced diminished text-to-image controllability. 
Pre-trained GLIGEN responded to stylistic text prompts (e.g., ``rustic wooden table''), but finetuned variants largely ignored such descriptors, producing uniform visual styles. 
This degradation is attributed to limited diversity in template-generated training captions (Figure~\ref{fig:generated2}).

\textbf{Trade-offs:} ControlNet preserves text-based controllability, enabling stylistic manipulation through prompt engineering, but suffers from higher hallucination rates. 
GLIGEN provides superior layout fidelity with minimal hallucination but sacrifices prompt-based stylistic control after finetuning. 
For applications prioritizing compositional accuracy, GLIGEN is preferable. 
For applications valuing stylistic flexibility, ControlNet may be more suitable.
See Figure~\ref{fig:generated_all} for examples of generated images with both approaches.

\subsection{End-to-End System Evaluation}

\begin{figure}[!htb]
    \centering
     \begin{subfigure}[b]{\textwidth}
         \centering
         \includegraphics[width=1\linewidth]{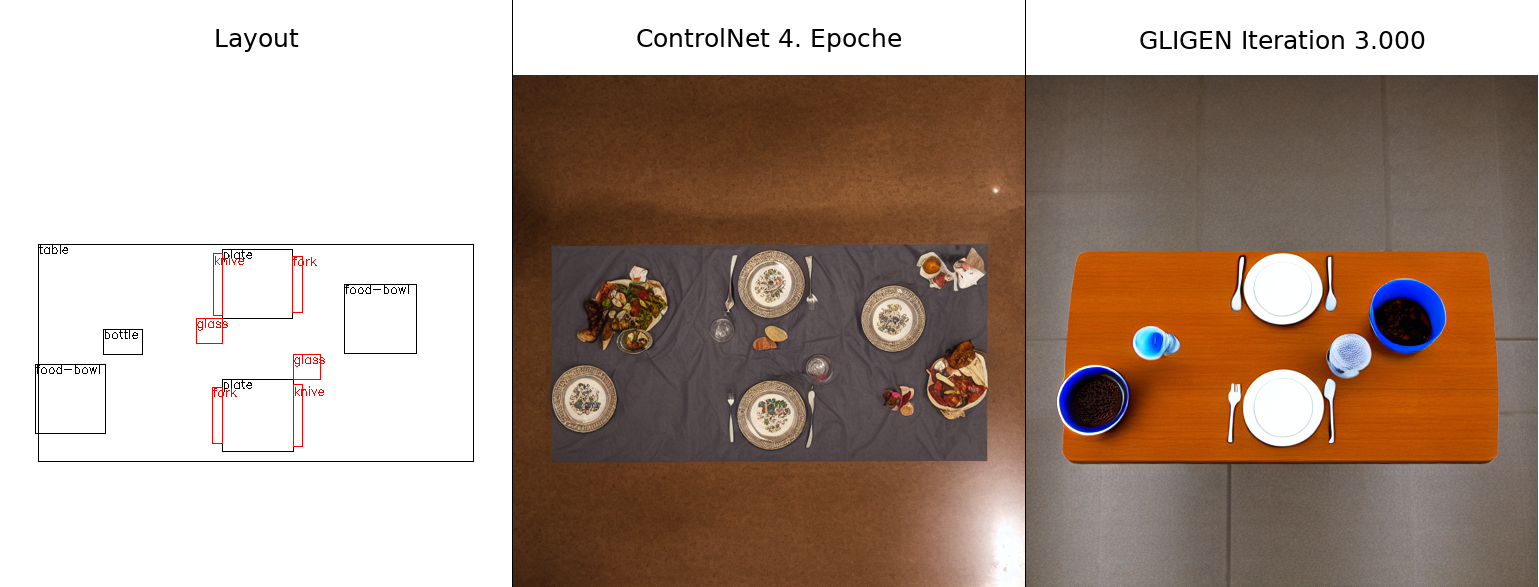}
         \caption{End-to-end generation for a two-person table setting.}
         \label{fig:generated_2p}
     \end{subfigure}
     \vspace{2em}
\begin{subfigure}[b]{\textwidth}
         \centering
         \includegraphics[width=\linewidth]{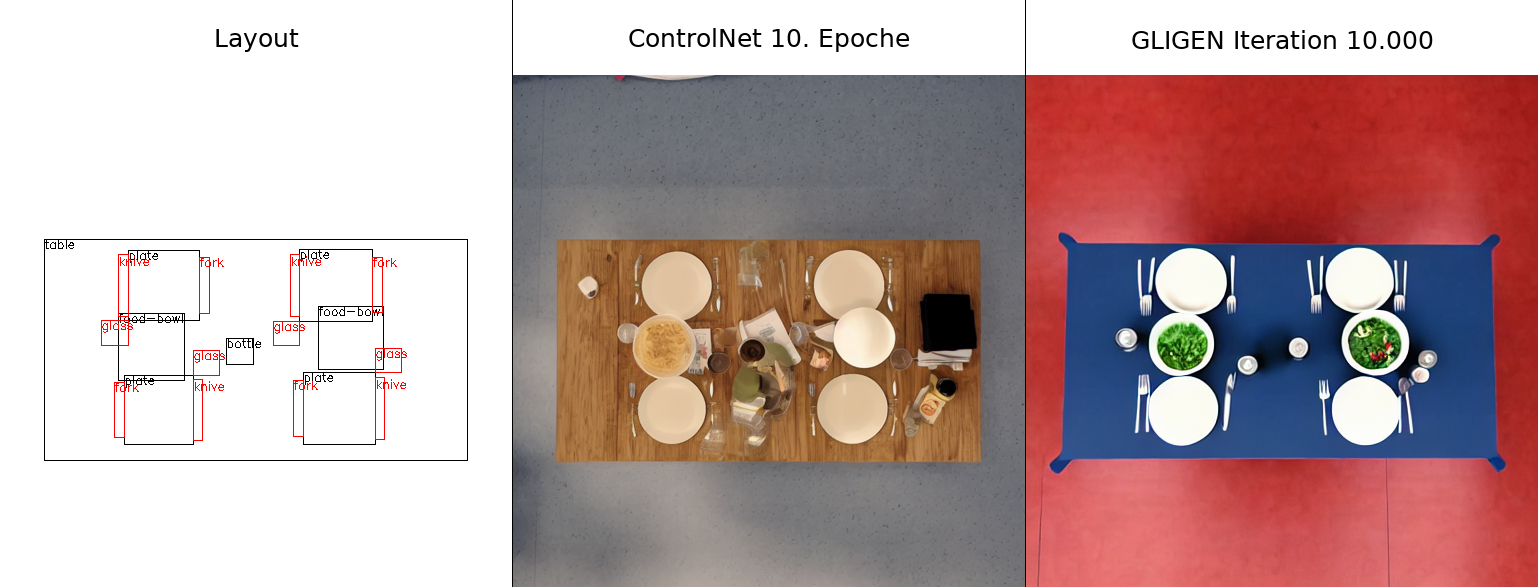}
         \caption{End-to-end generation for a four-person table setting.}
         \label{fig:generated_4p}
     \end{subfigure}
        \caption{End-to-end results of the complete system for two-person (\subref{fig:generated_2p}) and four-person (\subref{fig:generated_4p}) table settings. For each setting, a layout was first generated using the plates-only approach with rule-based completion. 
        The resulting layout was then used to generate images with both finetuned ControlNet and GLIGEN. 
        The system successfully produces images with correct object counts and plausible spatial arrangements, demonstrating the viability of the two-stage approach for compositionally controlled synthesis.
        }
        \label{fig:generated_2+4p}
\end{figure}

The complete pipeline—LLM-based layout generation, rule-based completion, quality scoring, and finetuned diffusion synthesis—was evaluated on held-out test cases. 
For each test case, five layout candidates were generated using \textit{Plates Only}, the highest-scoring layout was selected, completed via rules, and rendered using both ControlNet and GLIGEN (see Figure~\ref{fig:generated_2+4p}).

The system successfully generates images containing specified objects in plausible spatial arrangements. 
Plates position at table edges, cutlery flanks plates appropriately, glasses occupy conventional positions, and serving vessels are centrally located. 
The two-stage decomposition effectively overcomes compositional limitations of standard text-to-image models.

Limitations persist: occasional overlaps from completion propagate to synthesis, ControlNet hallucination introduces spurious objects, and GLIGEN's reduced text controllability limits stylistic variation. 
Nevertheless, the system demonstrates the viability of the two-stage paradigm for layout-controlled generation.

\section{Conclusion}
\label{sec:conclusion}

This paper presented a two-stage system for layout-controlled image generation that addresses a fundamental limitation of text-to-image diffusion models: the inability to generate images with precise compositional constraints. 
By decoupling spatial planning from visual synthesis, the approach leverages complementary strengths of Large Language Models for layout generation and diffusion models for photorealistic rendering.

The central finding is that task complexity critically impacts LLM-based spatial reasoning. 
Direct generation of complex layouts (15-25 objects) yielded poor recall (57.2\%), but task decomposition—generating simplified layouts with only core objects, followed by rule-based completion—dramatically improved performance to near-perfect recall (99.9\%) and precision (98.8\%). 
This demonstrates that current LLMs benefit substantially from explicit task simplification when confronted with high-dimensional spatial planning.

For image synthesis, comparative evaluation revealed critical trade-offs. 
ControlNet preserved text-based stylistic controllability but exhibited significant hallucination, generating objects absent from conditioning layouts. 
GLIGEN provided superior layout fidelity with minimal hallucination but lost responsiveness to stylistic text prompts after domain-specific finetuning. 
These findings indicate that architectural choices in conditioning mechanisms fundamentally influence the balance between compositional accuracy and generative flexibility.

\subsection{Limitations and Future Work}

Several limitations constrain the system's applicability. The rule-based completion algorithm relies on domain-specific conventions that may not generalize beyond table-setting. The layout quality scoring function employs manually tuned heuristics that may not comprehensively evaluate spatial plausibility across diverse contexts. 
Evaluation focused exclusively on top-down perspectives, potentially limiting robustness to oblique viewpoints. 
The observed ControlNet hallucination and GLIGEN text controllability loss represent unresolved challenges limiting practical deployment.

Future work should explore learned completion approaches through LLM fine-tuning or reinforcement learning, architectural modifications to reduce ControlNet hallucination while preserving flexibility, improved training strategies for GLIGEN using diverse captions to maintain prompt responsiveness, and extension to more complex spatial relationships beyond bounding boxes. 
Evaluation on diverse domains, including architectural design, retail arrangement, and UI layout generation would establish the generality of the two-stage paradigm.

Despite these limitations, this work demonstrates the viability of decomposing compositionally controlled image generation into explicit layout planning and layout-conditioned synthesis, successfully generating images with specified objects in plausible spatial arrangements and establishing a foundation for future research in compositionally constrained visual synthesis.

\subsubsection*{Acknowledgements} 
The research reported in this paper has been supported by the German Research Foundation DFG, as part of Collaborative Research Center (Sonderforschungsbereich) 1320 Project-ID 329551904 “EASE - Everyday Activity Science and Engineering”, University of Bremen (http://www.ease-crc.org/). 
The research was conducted in subprojects H01 "Sensorimotor and Causal Human Activity Models for Cognitive Architectures" and H03 "Discriminative and Generative Human Activity Models for Cognitive Architectures".
%
%
%
\bibliographystyle{splncs04}
\bibliography{references}

@inproceedings{Rombach2022,
  author    = {Rombach, Robin and Blattmann, Andreas and Lorenz, Dominik and Esser, Patrick and Ommer, Bj\"{o}rn},
  title     = {High-Resolution Image Synthesis with Latent Diffusion Models},
  booktitle = {Proceedings of the IEEE/CVF Conference on Computer Vision and Pattern Recognition (CVPR)},
  pages     = {10684--10695},
  year      = {2022}
}

@inproceedings{Ho2020,
  author    = {Ho, Jonathan and Jain, Ajay and Abbeel, Pieter},
  title     = {Denoising Diffusion Probabilistic Models},
  booktitle = {Advances in Neural Information Processing Systems (NeurIPS)},
  volume    = {33},
  pages     = {6840--6851},
  year      = {2020}
}

@article{HoSalimans2022,
  author    = {Ho, Jonathan and Salimans, Tim},
  title     = {Classifier-Free Diffusion Guidance},
  journal   = {arXiv preprint arXiv:2207.12598},
  year      = {2022}
}

@inproceedings{Zhang2023,
  author    = {Zhang, Lvmin and Rao, Anyi and Agrawala, Maneesh},
  title     = {Adding Conditional Control to Text-to-Image Diffusion Models},
  booktitle = {Proceedings of the IEEE/CVF International Conference on Computer Vision (ICCV)},
  pages     = {3836--3847},
  year      = {2023}
}

@inproceedings{Li2023,
  author    = {Li, Yuheng and Liu, Haotian and Wu, Qingyang and Mu, Fangzhou and Yang, Jianwei and Gao, Jianfeng and Li, Chunyuan and Lee, Yong Jae},
  title     = {{GLIGEN}: Open-Set Grounded Text-to-Image Generation},
  booktitle = {Proceedings of the IEEE/CVF Conference on Computer Vision and Pattern Recognition (CVPR)},
  pages     = {22511--22521},
  year      = {2023}
}

@inproceedings{Feng2023,
  author    = {Feng, Weixi and Zhu, Wanrong and Fu, Tsu-Jui and Jampani, Varun and Akula, Arjun and He, Xuehai and Basu, Sugato and Wang, Xin Eric and Wang, William Yang},
  title     = {{LayoutGPT}: Compositional Visual Planning and Generation with Large Language Models},
  booktitle = {Advances in Neural Information Processing Systems (NeurIPS)},
  volume    = {36},
  year      = {2023}
}

@article{Lian2023,
  author    = {Lian, Long and Li, Boyi and Yala, Adam and Darrell, Trevor},
  title     = {{LLM-grounded Diffusion}: Enhancing Prompt Understanding of Text-to-Image Diffusion Models with Large Language Models},
  journal   = {Transactions on Machine Learning Research (TMLR)},
  year      = {2024},
  issn      = {2835-8856}
}

@inproceedings{Jampour2017,
  author    = {Jampour, Mahyar and Dana, Kristin and Rom, Hillel and Jojic, Nebojsa},
  title     = {Table-Setting: A Dataset for Object-Based Reasoning in a Visually Rich Domain},
  booktitle = {Proceedings of the IEEE Conference on Computer Vision and Pattern Recognition Workshops (CVPRW)},
  pages     = {1265--1274},
  year      = {2017}
}

@inproceedings{Kirillov2023,
  author    = {Kirillov, Alexander and Mintun, Eric and Ravi, Nikhila and Mao, Hanzi and Rolland, Chloe and Gustafson, Laura and Xiao, Tete and Whitehead, Spencer and Berg, Alexander C. and Lo, Wan-Yen and Dollár, Piotr and Girshick, Ross},
  title     = {Segment Anything},
  booktitle = {Proceedings of the IEEE/CVF International Conference on Computer Vision (ICCV)},
  pages     = {3879--3890},
  year      = {2023}
}

@inproceedings{Bao2022,
  author    = {Bao, Hangbo and Dong, Li and Piao, Songhao and Wei, Furu},
  title     = {{BEiT}: BERT Pre-Training of Image Transformers},
  booktitle = {International Conference on Learning Representations (ICLR)},
  year      = {2022}
}

@inproceedings{Ramesh2022,
  author    = {Ramesh, Aditya and Dhariwal, Prafulla and Nichol, Alex and Chu, Casey and Chen, Mark},
  title     = {Hierarchical Text-Conditional Image Generation with {CLIP} Latents},
  journal   = {arXiv preprint arXiv:2204.06125},
  year      = {2022}
}

@inproceedings{Saharia2022,
  author    = {Saharia, Chitwan and Chan, William and Saxena, Saurabh and Li, Lala and Whang, Jay and Denton, Emily and Ghasemipour, Kamyar Seyed and Ayan, Burcu Karagol and Mahdavi, S. Sara and Gontijo Lopes, Rapha and Salimans, Tim and Ho, Jonathan and Fleet, David J. and Norouzi, Mohammad},
  title     = {Photorealistic Text-to-Image Diffusion Models with Deep Language Understanding},
  booktitle = {Advances in Neural Information Processing Systems (NeurIPS)},
  volume    = {35},
  pages     = {36479--36494},
  year      = {2022}
}

@inproceedings{Chefer2023,
  author    = {Chefer, Hila and Schwartz, Yuval and Lev-Ari, Tomer and Wolf, Lior},
  title     = {Attend-and-Excite: Attention-Based Semantic Guidance for Text-to-Image Diffusion Models},
  booktitle = {ACM SIGGRAPH 2023 Conference Proceedings},
  year      = {2023}
}

@inproceedings{Song2021,
    title={Denoising Diffusion Implicit Models},
    author={Song, Jiaming and Meng, Chenlin and Ermon, Stefano},
    booktitle={International Conference on Learning Representations (ICLR)},
    year={2021}
}

@article{Cho2023,
  author    = {Cho, Jaemin and Zala, Abhay and Bansal, Mohit},
  title     = {Visual Programming for Text-to-Image Generation and Evaluation},
  journal   = {arXiv preprint arXiv:2305.13655},
  year      = {2023}
}

@article{Tang2023,
  author    = {Tang, Zecheng and Wu, Chenfei and Li, Juntao and Duan, Nan},
  title     = {{LayoutNUWA}: Revealing the Hidden Layout Expertise of Large Language Models},
  journal   = {arXiv preprint arXiv:2311.16015},
  year      = {2023}
}

@inproceedings{Li2019,
  author    = {Li, Jianan and Yang, Jimei and Hertzmann, Aaron and Zhang, Jianming and Xu, Tingfa},
  title     = {LayoutGAN: Generating Graphic Layouts with Wireframe Discriminators},
  booktitle = {Proceedings of the 27th ACM International Conference on Multimedia},
  pages     = {1363--1371},
  year      = {2019}
}

@inproceedings{Jyothi2019,
  author    = {Jyothi, Akash Abdu and Durand, Thibaut and He, Jiawei and Sigal, Leonid and Mori, Greg},
  title     = {LayoutVAE: Stochastic Scene Layout Generation from a Label Set},
  booktitle = {Proceedings of the IEEE/CVF Conference on Computer Vision and Pattern Recognition (CVPR)},
  pages     = {11825--11833},
  year      = {2019}
}

@inproceedings{Gupta2020,
  author    = {Gupta, Kamal and Lazarow, Justin and Achille, Alessandro and Davis, Larry S. and Mahadevan, Vijay and Shrivastava, Abhinav},
  title     = {LayoutTransformer: Layout Generation and Completion with Self-attention},
  booktitle = {European Conference on Computer Vision (ECCV)},
  pages     = {273--289},
  year      = {2020}
}

@inproceedings{Arroyo2021,
  author    = {Arroyo, Diego Martin and Postels, Janis and Tombari, Federico},
  title     = {Variational Transformer Networks for Layout Generation},
  booktitle = {British Machine Vision Conference (BMVC)},
  year      = {2021}
}

@article{Brown2020,
  author    = {Brown, Tom B. and Mann, Benjamin and Ryder, Nick and Subbiah, Melanie and Kaplan, Jared D. and Dhariwal, Prafulla and Neelakantan, Arvind and Shyam, Pranav and Sastry, Girish and Askell, Amanda and Agarwal, Sandhini and Herbert-Voss, Ariel and Krueger, Gretchen and Henighan, Tom and Child, Rewon and Ramesh, Aditya and Ziegler, Daniel M. and Wu, Jeffrey and Winter, Clemens and Hesse, Christopher and Chen, Mark and Sigler, Eric and Litwin, Mateusz and Gray, Scott and Chess, Benjamin and Clark, Jack and Berner, Christopher and McCandlish, Sam and Radford, Alec and Sutskever, Ilya and Amodei, Dario},
  title     = {Language Models are Few-Shot Learners},
  journal   = {arXiv preprint arXiv:2005.14165},
  year      = {2020}
}

@article{OpenAI2023,
  author    = {{OpenAI}},
  title     = {GPT-4 Technical Report},
  journal   = {arXiv preprint arXiv:2303.08774},
  year      = {2023}
}

@INPROCEEDINGS{TSD2025,
	author={C. {Mason} and K. {Gadzicki} and M. {Meier} and F. {Ahrens} and T. {Kluss} and J. {Maldonado} and F. {Putze} and T. {Fehr} and C. {Zetzsche} and M. {Herrmann} and K. {Schill} and T. {Schultz}},
	booktitle={2020 IEEE/RSJ International Conference on Intelligent Robots and Systems (IROS)}, 
	title={From Human to Robot Everyday Activity}, 
	year={2020},
	pages={8997-9004},
	doi={10.1109/IROS45743.2020.9340706},
}
\end{document}